\pdfoutput=1

\documentclass[11pt]{article}

\usepackage{ACL2023}

\usepackage{times}
\usepackage{latexsym}

\usepackage[T1]{fontenc}

\usepackage{color}
\usepackage{array}
\usepackage{enumitem}
\usepackage{caption}
\usepackage{algorithm}
\usepackage{algorithmic}
\usepackage{amsmath}
\usepackage{amssymb}
\usepackage{graphicx}
\usepackage{multirow}
\usepackage{booktabs}
\usepackage{arydshln}
\usepackage{bm}
\usepackage{tikz}
\usepackage{pgfplots}
\usepackage{xspace}
\usepackage{paralist}
\usepackage{tabularx}
\usepackage{graphicx}
\usepackage{subcaption}

\usepackage[utf8]{inputenc}

\usepackage{microtype}

\usepackage{inconsolata}

\usepackage{color,xcolor}
\usepackage{colortbl}
\usepackage{soul}

\usepackage{tcolorbox}

\usepackage{adjustbox}

  {\list{}{\leftmargin=0.3in\rightmargin=0.3in}\item[]}%
  {\endlist}

\definecolor{nmgray}{RGB}{229,229,229}
\definecolor{underlinegray}{RGB}{197,197,197}

\definecolor{introblue}{RGB}{0,176,240}
\definecolor{introgreen}{RGB}{0,203,134}
\definecolor{introgreen2}{RGB}{139,243,206}

\setul{0.3ex}{0.2ex}
\setulcolor{underlinegray}

\newtcolorbox{mybox}[2][]{
width=\columnwidth,
colback = nmgray!75!white, 
colframe = nmgray!75!white, 
boxsep=0pt,left=10pt,right=10pt,top=0pt,bottom=0pt,
fontupper=\linespread{0.9}\selectfont,
title=#2,#1}

\newcommand{\mlogo}{\raisebox{-6pt}{\includegraphics[width=1.6em]{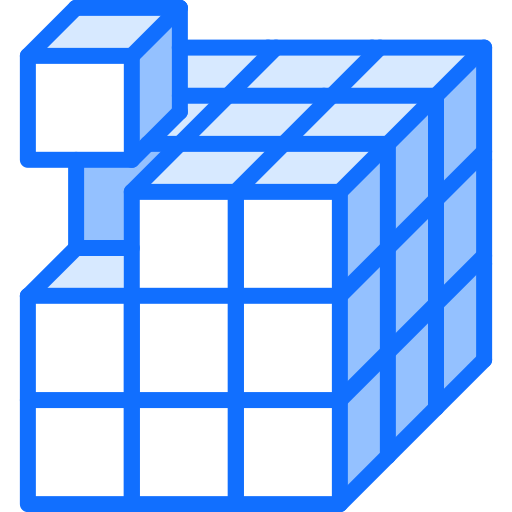}}\xspace}

\title{\mlogo\texttt{XNLP}: An Interactive Demonstration System\\ for Universal Structured NLP}

\author{
Hao Fei\textsuperscript{\rm 1},  \quad
Meishan Zhang\textsuperscript{\rm 2},  \quad
Min Zhang\textsuperscript{\rm 2},  \quad
Tat-Seng Chua\textsuperscript{\rm 1} \\
\textsuperscript{\rm 1} NExT++ Research Center, National University of Singapore \\
\textsuperscript{\rm 2} Harbin Institute of Technology (Shenzhen), China \\
\tt {\{haofei37,dcscts\}@nus.edu.sg, \{zhangmeishan,zhangmin2021\}@hit.edu.cn}
}

\begin{document}
\maketitle
\begin{abstract}
Structured Natural Language Processing (XNLP) is an important subset of NLP that entails understanding the underlying semantic or syntactic structure of texts, which serves as a foundational component for many downstream applications.
Despite certain recent efforts to explore universal solutions for specific categories of XNLP tasks, a comprehensive and effective approach for unifying all XNLP tasks long remains underdeveloped.
Meanwhile, while XNLP demonstration systems are vital for researchers exploring various XNLP tasks, existing platforms can be limited to, e.g., supporting few XNLP tasks, lacking interactivity and universalness. 
To this end, we propose an advanced XNLP demonstration system, where we leverage LLM to achieve universal XNLP, \emph{with one model for all} with high generalizability.
Overall, our system advances in multiple aspects, including universal XNLP modeling, high performance, interpretability, scalability, and interactivity, offering a unified platform for exploring diverse XNLP tasks in the community.\footnote{\texttt{XNLP} is online: \url{https://xnlp.haofei.vip}\\ Video demonstration at \url{https://youtu.be/bOc-9HELEVw}}
\end{abstract}

\section{Introduction}
\begin{figure}[!t]
  \centering
  \includegraphics[width=0.98\columnwidth]{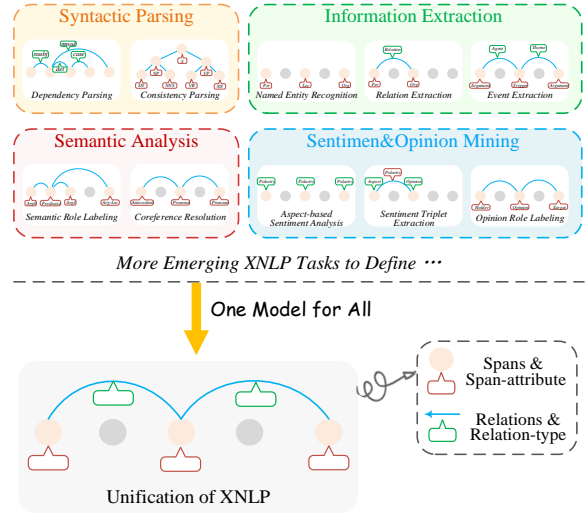}
  \caption{
  Illustration of the Structured NLP (XNLP) tasks, and the unification of XNLP by decomposing into the predictions of spans and relations.
  }
  \label{fig:intro}
  \vspace{-10pt}
\end{figure}

XNLP has been referred to as a special form of NLP tasks that involves holistically analyzing and interpreting the underlying semantic or syntactic structure within a text, such as 
Syntactic Dependency Parsing \cite{nivre-2003-efficient}, 
Information Extraction \cite{wang-cohen-2015-joint}, 
Coreference Resolution \cite{lee-etal-2017-end}, 
and Opinion Extraction \cite{pontiki-etal-2016-semeval}, etc.
Figure \ref{fig:intro} (upper part) illustrates some representative XNLP tasks under different categories.
XNLP has been infrastructural for a wide range of downstream NLP applications, such as Knowledge Graph Construction \cite{bosselut-etal-2019-comet}, Empathetic Dialogue \cite{rashkin-etal-2019-towards}, and more newly-emerging applications and techniques \cite{tang-etal-2020-dependency}.

As the key common characteristics, all the XNLP tasks have revolved around predicting two key elements from input: \textbf{1) textual spans} and \textbf{2) relations between spans} \cite{FeiLasuieNIPS22}, as depicted in the Figure \ref{fig:intro} (lower part).
Traditional efforts for XNLP have treated each task independently, which has led to limited utilization of shared features among XNLP tasks \cite{he-etal-2019-interactive}, and sub-optimal model generalization across different datasets \cite{chauhan-etal-2020-sentiment}, such as cross-language and cross-domain scenarios.
In this paper, we emphasize the importance of Model Unification as a crucial topic in NLP. By unifying various NLP tasks under the XNLP framework, we can take advantage of the shared characteristics among tasks, leading to better model generalization and improved performance in realistic scenarios of product deployment.

Despite recent certain efforts in exploring universal solutions for some categories of XNLP tasks, such as Unified Sentiment Analysis \cite{chen-qian-2020-relation,0001LLWLJ22}, and Universal Information Extraction \citep[UIE;][]{lu-etal-2022-unified,FeiLasuieNIPS22}, a comprehensive and effective approach for unifying all XNLP tasks has not been fully established.
Fortunately, Large Language Models (LLMs) \cite{VaswaniSPUJGKP17,RaffelSRLNMZLL20} present a potential solution for unification across all XNLP tasks.
There is a recent development in the form of LLMs, e.g., ChatGPT \cite{ouyang2022training}, LLaMA \cite{abs-2302-13971} and Vicuna \cite{abs-2304-03277}, that have shown promising advancements in NLP and other fields.
LLMs, with sufficient sizes of model and data, have demonstrated impressive generalization capabilities, well supporting the idea of \emph{``One model for all''} \cite{abs-2303-08774}. 
In this work, we propose taking advantage of LLMs to achieve universal XNLP, addressing the lack of a well-defined and holistic approach.

On the other hand, demonstration systems play a crucial role for researchers (especially beginners) exploring various XNLP tasks, providing a platform to analyze and understand the functionalities of different NLP components and their applications.
While there are existing widely-used XNLP demo systems, such as 
\emph{CoreNLP}\footnote{\url{http://corenlp.run/}}, 
\emph{AllenNLP}\footnote{\url{https://demo.allennlp.org/}}, 
we have observed several key issues with them:
1) limited to only a few specific tasks;
2) lacking interactive and extensible features, making it challenging to support dynamic growth in new XNLP tasks;
3) not universal systems, requiring separate models for each task, which can lead to increased overhead.
To address these limitations, this work aims to build an advanced platform that provides superior XNLP demonstrations and benefits the broader NLP community.

In summary, our system advances in the following aspects.

\setdefaultleftmargin{1.5em}{1.0em}{1.87em}{1.7em}{1em}{1em}
\begin{compactenum}
  \item[1)] \textbf{Universalness} 
    \setdefaultleftmargin{0em}{1.0em}{1.87em}{1.7em}{1em}{1em}
    \begin{compactitem}
        \item Our XNLP system takes the existing open-source LLMs as the backbone engine with excellent generalization capabilities, enabling unified prediction of various XNLP tasks, leading to a streamlined and cohesive XNLP ecosystem.
  
        \item The LLM-based system supports end-to-end predictions for complex structured tasks, regardless of whether the spans are nested or discontinuous, making it versatile and adaptable to different linguistic structures.
  
    \end{compactitem}

  \item[2)] \textbf{High Performance} 
    \begin{compactitem}

        \item Our system is capable of few-shot or weakly-supervised learning. Having undergone extensive pre-training, LLMs do not require in-domain fine-tuning on specific task data.

        \item Our system supports open-label and vocabulary predictions, utilizing LLM's generalization capabilities to discover new labels and vocabs with superior out-of-domain generalization.
  
        \item Our approach naturally lends itself to cross-lingual, code-switching, and cross-domain settings.
  
    \end{compactitem}

  \item[3)] \textbf{Scalability\&Interpretability\&Interactivity} 
    \begin{compactitem}

        \item The system allows dynamic addition and definition of new tasks, requiring users only to provide demonstrations for the new tasks.

        \item Predictions generated by our system are interpretable, as LLMs are able to provide rationales for their decisions, explaining why a specific result is produced.
  
        \item The system enables user-machine interaction, empowering users to provide feedback, thereby allowing the system to refine its predictions based on user input.

    \end{compactitem}

\end{compactenum}

\section{Related Work}

\vspace{-1.5mm}
\subsection{Structured NLP}

\vspace{-2mm}
Over the last few decades, XNLP has garnered significant research attention, with several works addressing specific aspects of XNLP tasks, spanning from linguistic/syntactic parsing \cite{kitaev2018constituency}, to information extraction \cite{mikheev1999named}, to semantic analysis \cite{he2017deep} and to sentiment analysis \& opinion mining \cite{Wu0RJL21}.
Prior studies and efforts have been paid and achieved notable developments for each of the XNLP tasks, such as Syntactic Dependency Parsing \cite{nivre-2003-efficient}, 
Information Extraction \cite{wang-cohen-2015-joint}, 
Coreference Resolution \cite{lee-etal-2017-end}, 
and Opinion Extraction \cite{pontiki-etal-2016-semeval}, etc.
Different XNLP tasks may have different specific task definitions, while prediction formats of all the XNLP tasks can be reduced to the same prototype: the term extraction and relation detection \cite{lu-etal-2022-unified,FeiLasuieNIPS22}.

\vspace{-2mm}
\paragraph{Demonstration for XNLP.}
The development of demonstration platforms has been crucial for educational and academic purposes, e.g., aiding researchers to explore various tasks and gaining hands-on experiences.
Existing widely-employed open demo systems for XNLP include \emph{CoreNLP}\footnote{\url{http://corenlp.run/}}, \emph{AllenNLP}\footnote{\url{https://demo.allennlp.org/}} and \emph{Explosion.ai}\footnote{\url{https://explosion.ai/}} etc.
While offering user-friendly web interface for users to access a set of XNLP functionalities, 
there remain certain limitations, such as 
lacking flexibility for incorporating new tasks, non-universalness for model and cross-domain generalization.

\begin{figure}[!t]
  \centering
  \includegraphics[width=1\columnwidth]{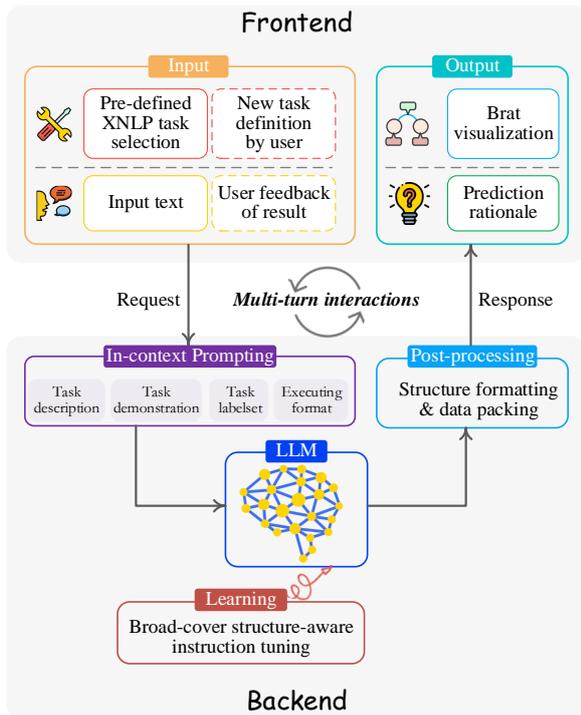}
  \caption{
  The overall architecture of our \texttt{XNLP} system includes the \textbf{frontend} module and the \textbf{backend} module.
  }
  \label{fig:framework}
  \vspace{-10pt}
\end{figure}

\vspace{-2mm}
\subsection{Model Unification}

There have been notable efforts to explore universal modeling for a type of NLP tasks \cite{chen-qian-2020-relation,0001LLWLJ22,lu-etal-2022-unified,FeiLasuieNIPS22}, showcasing the benefit and potential of model unification, e.g., better leverage of shared characteristics and knowledge across tasks, simplified model maintenance, and enhanced system efficiency.
However, a comprehensive and effective approach for unifying all XNLP tasks remains under-investigated. 
In this work, by capitalizing on LLM's robustness and broad applicability, we aim to pave the way for an advanced unified framework capable of handling diverse XNLP tasks effectively.

\vspace{-2mm}

\section{System Design}

\vspace{-2mm}
\paragraph{Architecture overview.}

We design our \texttt{XNLP} demo system into a web interface form.
Built based on the Django\footnote{\url{https://www.djangoproject.com/}, v4.2.4} framework, \texttt{XNLP} divides the functions into the \textbf{frontend} module and the \textbf{backend} module.
As shown in Figure \ref{fig:framework}, the frontend takes user inputs and displays the visualization of outputs, and the backend provides task prediction services with LLM as its core engine, based on the in-context learning paradigm.
Also, it is possible for multi-turn interactions between frontend and backend.

\vspace{-2mm}

\subsection{Backend}

\paragraph{Backbone LLM.}

Among a list of open-source LLMs, we consider the Vicuna-13B\footnote{\url{https://github.com/lm-sys/FastChat}} as our backbone.
Trained by fine-tuning LLaMA \cite{abs-2302-13971} on user-shared conversations collected from ShareGPT, Vicuna has achieved more than 90\% of OpenAI ChatGPT's \cite{abs-2304-03277} quality in user preference tests.

\vspace{-2mm}
\paragraph{In-context learning.}
To elicit LLM to induce task predictions, we build in-context prompts.
We note that to ensure the support of universal XNLP for any potential tasks and inputs, the prompt template should cover rich and informative information from the user end.
Thus, we design the prompt by mainly covering the  \textbf{task name}, \textbf{task description}, \textbf{task demonstration}, \textbf{task label set}, \textbf{executing format}, \textbf{input text}, \textbf{language} and \textbf{domain}.

\vspace{-3pt}
\begin{mybox}\texttt
\texttt{\{Task-desc\}}\\
\texttt{----------------------------------------------------------------------------------------------------------}\\
\texttt{For example, \{Task-demo\}}\\
\texttt{----------------------------------------------------------------------------------------------------------}\\
\texttt{Note the task output format should be with this: \{Exe-format\}}\\
\texttt{----------------------------------------------------------------------------------------------------------}\\
\texttt{And generally, the desired predicted labels should be within the following given label set: \{Task-label\}}\\
\texttt{----------------------------------------------------------------------------------------------------------}\\
\texttt{Now, given a new test input: ``\{Input-text\}'', please do the task of \{Task-name\}.}\\
\texttt{----------------------------------------------------------------------------------------------------------}\\
\texttt{Note the input is with \{Language\} language, and the text is from the \{Domain\} domain.}\\
\texttt{----------------------------------------------------------------------------------------------------------}\\
\texttt{Please predict all possible results strictly following the exact given format, without any other output of explanations.}
\end{mybox}
\vspace{-3pt}
Fed with the above prompt, the LLM is expected to output prediction in the provided format (\textbf{executing format}), with which, the post-process program further parses and polishes the structure result, and packs data to return to the frontend.

\begin{figure*}[!t]
\centering
\includegraphics[width=1\textwidth]{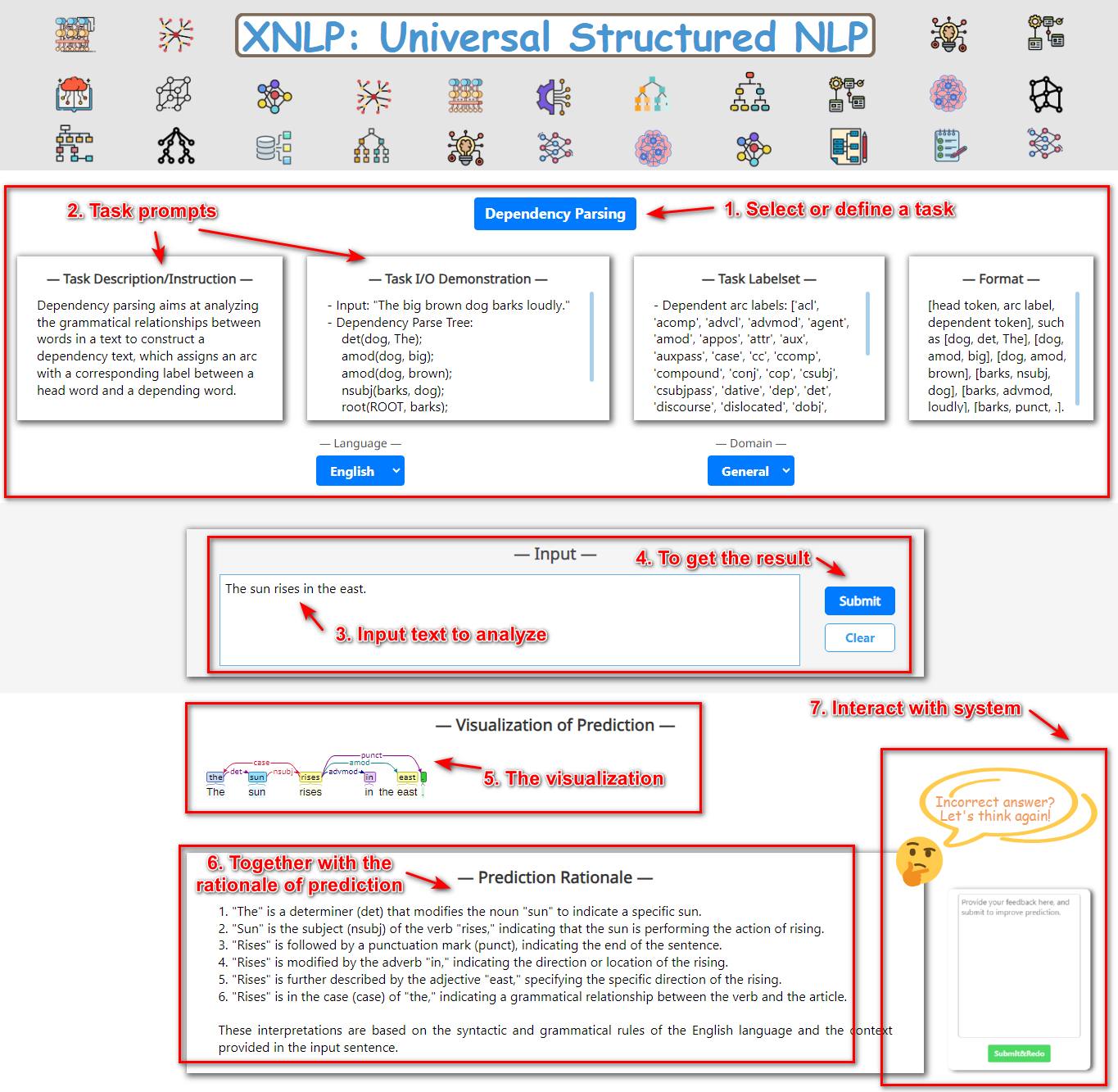}
\caption{
Screenshot of the \texttt{XNLP} web application, where key functions are annotated.
}
\label{fig:overview}
\vspace{-10pt}
\end{figure*}

\vspace{-2mm}
\paragraph{Broad-cover structure-aware instruction tuning.}

While LLM's outputs are sequential, XNLP tasks are highly structured.
Thus, we expect the LLM to generate strictly structural results conditioned on sequence inputs.
We consider further tuning the LLM with a \emph{broad-cover structure-aware instruction tuning} mechanism.
Instruction tuning is an emergent paradigm of LLM fine-tuning wherein natural language instructions are leveraged with LLM to induce the desired result more accurately.
We write the XNLP predictions (outputs) for any input prompt by formatting the predictions into task-agnostic (i.e., task broad-covering) well-formed structure representations as in \citet{FeiLasuieNIPS22}.

\begin{figure}[!t]
\centering
\includegraphics[width=0.98\columnwidth]{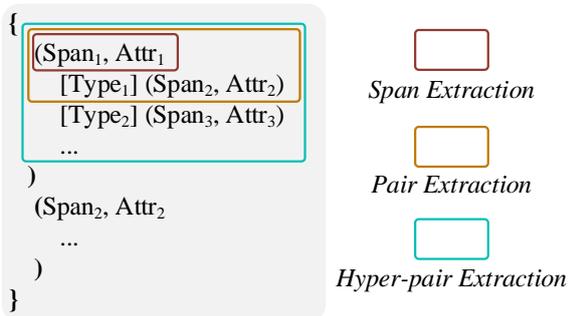}
\vspace{-2mm}
\caption{
Structure formatter for universal XNLP.
}
\label{fig:formatter}
\vspace{-14pt}
\end{figure}

\begin{figure*}[!t]
\centering
\includegraphics[width=1\textwidth]{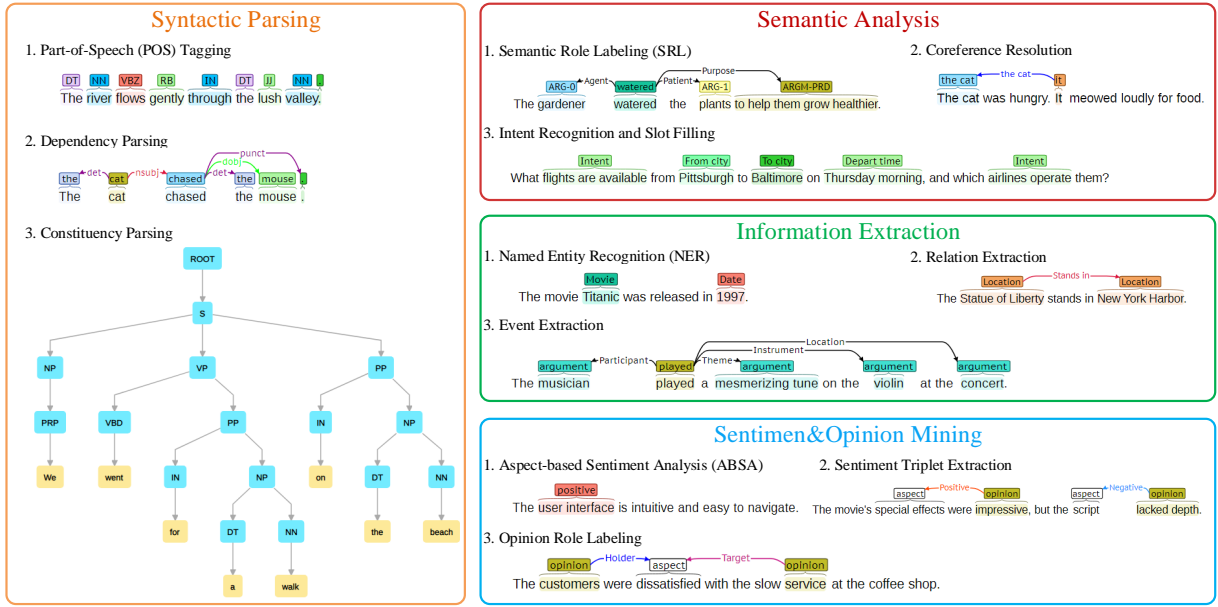}
\caption{
Screenshots of the visualizations of 12 representative XNLP tasks.
Best viewing with zooming in.
}
\label{fig:vis-all}
\vspace{-10pt}
\end{figure*}

\vspace{-3mm}
\paragraph{Structure formatting.}

As aforementioned, all the XNLP can be unified by predicting two key elements: the term extraction (with the span attribute) and relation detection (with the relation type), as illustrated in Figure \ref{fig:intro}.
To unify all XNLP tasks, we follow \citet{FeiLasuieNIPS22} and design a structure formatter, where all the XNLP task outputs share the same structural representations.
As shown in Figure \ref{fig:formatter}, under the structure formatted, all XNLP tasks have been divided into the span extraction, pair extraction and hyper-pair extraction.

\subsection{Frontend}

As illustrated in Figure \ref{fig:framework} (upper part), the frontend of \texttt{XNLP} receives inputs of 1) texts or user feedback or 2) task metadata (pre-defined or user-defined), and exhibits outputs from LLM.
Following we mainly describe the key features of the frontend module as listed below.

\vspace{-2mm}
\paragraph{Pre-defined XNLP tasks.}
To facilitate the user operation, we pre-defined total 22 XNLP tasks, covering four frequent categories, including Syntax Parsing, Information Extraction, Semantic Analysis and Sentiment/Opinion Mining.

\vspace{-1mm}
\paragraph{New task definition.}
As there are rapidly-emergent XNLP tasks in the NLP community, it is impossible to cover it all in the pre-definition.
We thus allow users to define their own XNLP tasks.
This can be easily accomplished in our system without much effort, as the LLM has exceptional zero-shot performance and understanding ability.
We require from the user only the \emph{task name}, \emph{task description}, \emph{task demonstration}, \emph{task label set}, \emph{executing format}.

\vspace{-1mm}
\paragraph{XNLP structure visualization.}
The key role of \texttt{XNLP} system is the visualization of the task output structure.
We employ the open-source \emph{brat} system\footnote{\url{https://brat.nlplab.org/}} to realize this.
\emph{brat} has been shown very popular and effective in rendering structured data, with pretty visualization and stable functions.

\vspace{-1mm}
\paragraph{Rationale for explainable task prediction.}
Besides the visualization of direct task results, we also display the rationale for each prediction, allowing \emph{seeing what and knowing why}.
This is especially meaningful for the beginners of the researchers for XNLP tasks.
To enable this, we just ask LLM \emph{``How and why do you make your decision?''} after each task prediction.

\vspace{-1mm}
\paragraph{Enhancing prediction with user interaction.}

To take full advantage of the LLM, we further allow users to interact with our system by providing any feedbacks, so that users can revise the task predictions whenever they feel the results are not incorrect or coincident with their minds.
To reach this, we also add another round of query to LLM, by asking \emph{``The above prediction is not all right, because {Feedback}.
Please do the task again by carefully taking the feedback here''}.

\vspace{-1mm}
\section{System Walkthrough}

\vspace{-1mm}
Figure \ref{fig:overview} gives a comprehensive walkthrough of how the system can be operated by users.

\setdefaultleftmargin{5em}{2em}{}{}{}{}
\begin{compactenum}

    \item[$\blacktriangleright$ \textbf{Step-1.}] \emph{users select or define a task};

    \item[$\blacktriangleright$ \textbf{Step-2.}] \emph{users go through (for pre-defined) or fill in (for user-defined) the task prompt};

    \item[$\blacktriangleright$ \textbf{Step-3.}] \emph{users key in the text to analyze};

    \item[$\blacktriangleright$ \textbf{Step-4.}] \emph{users submit the text \& metadata and request result};

    \item[$\blacktriangleright$ \textbf{Step-5.}] \emph{users can browse the visualization of task output};

    \item[$\blacktriangleright$ \textbf{Step-6.}] \emph{users observe the rationale of this result};

    \item[$\blacktriangleright$ \textbf{Step-7.}] \emph{users can further provide feedback for the system to re-generate result};

\end{compactenum}

Following we demonstrate \texttt{XNLP} system by walking readers through several important functions.

\subsection{User-allowed Operations}

\paragraph{Pre-defined XNLP task selection.}
For the first step, users should select an XNLP task template from the 22 system pre-defined pools.
The operation is shown in Figure \ref{task-selection} in Appendix $\S$\ref{Selection of Pre-defined XNLP Tasks}.

\vspace{-1mm}
\paragraph{New task definition.}
Or, user can define their own tasks.
As shown in Figure \ref{fig:definition} in Appendix $\S$\ref{New XNLP Task Definition}, users should decide the task name, and fill in the task metadata (task prompt) as shown in Figure \ref{fig:overview}, and also select a pre-defined task with which the new task shares most similarity.

\vspace{-1mm}
\paragraph{Language and domain notifications.}

To enable more accurate predictions, it is better to explicitly notify LLM what language and domain the input has.
Figure \ref{fig:Chinese} in Appendix $\S$\ref{Text in Different Language} and Figure \ref{fig:domain} in Appendix $\S$\ref{Text in Different Domain} illustrate the operations, respectively.

\vspace{-1mm}
\paragraph{Improving/Revising Prediction with User Feedback}
Figure \ref{fig:interaction} in Appendix $\S$\ref{Multi-turn Interactions with User Feedback} showcase the operation for the multi-turn user interaction.

\vspace{-1mm}
\subsection{Task Visualization}

\vspace{-1mm}
Here we showcase the XNLP task visualizations of real examples via our system.
Figure \ref{fig:vis-all} renders the outputs for the four task clusters, with each showing three representative task results, such as:

\vspace{-2mm}
\paragraph{Syntax parsing,}
including Part-of-Speech (POS) Tagging, Dependency Parsing and Constituency Parsing.

\vspace{-2mm}
\paragraph{Semantic analysis,}
including Semantic Role Labeling (SRL), Coreference Resolution, and Intent Recognition and Slot Filling.

\vspace{-2mm}
\paragraph{Information extraction,}
including Named Entity Recognition (NER),  Relation Extraction, and Event Extraction.

\vspace{-2mm}
\paragraph{Sentiment/opinion mining,}
including Aspect-based Sentiment Analysis (ABSA), Sentiment Triplet Extraction and Opinion Role Labeling.

We can observe from the visualizations that, 
1) the structure visualizations are pretty, owing to the use of the brat system;
2) the results of tasks are correct, for which we give the credit to the integration of LLM, and also the \emph{broad-cover structure-aware instruction tuning} mechanism.

\begin{figure}[!t]
\centering
\includegraphics[width=0.98\columnwidth]{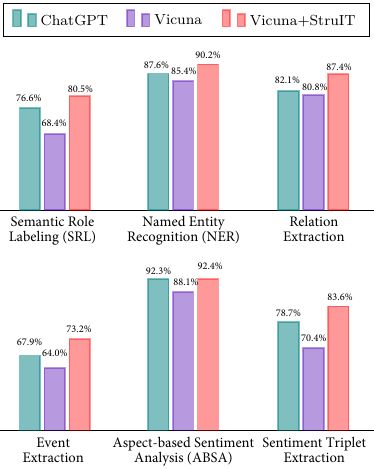}
\caption{
Comparisons (end-to-end prediction, in accuracy) between ChatGPT and Vicuna on XNLP tasks.
}
\label{performance}
\vspace{-10pt}
\end{figure}

\section{Performance Evaluation}

\vspace{-1mm}
To quantitatively verify the performance of the backbone LLM on XNLP tasks, we now perform evaluations.
We compare the Vicuna (13B) with the ChatGPT over 100 randomly selected test instances of 6 XNLP tasks.
The experiments are based on one-shot in-context learning, i.e., with one demonstration as input.
Figure \ref{performance} shows the comparisons.
We see Vicuna has a slightly lower performance than ChatGPT, while Vicuna after broad-cover structure-aware instruction tuning (Vicuna+StruIT) shows results even much better than ChatGPT, with smaller model size (13B vs. 175B).

\vspace{-1mm}
\section{Conclusion}

\vspace{-2mm}
We present \texttt{XNLP}, an advanced online demonstration system for interaction and visualization of XNLP tasks.
\texttt{XNLP}, built upon LLM, effectively models all the XNLP tasks universally, achieving \emph{one model for all} in zero-shot or weak supervision.
\texttt{XNLP} not only renders the output structures with delicate visualizations, but also provides rationales for interpretable predictions.
Also, \texttt{XNLP} allows the users to define newly emergent XNLP tasks; and enables users to dynamically revise the output with multi-turn interactions.
Our \texttt{XNLP} contributes to the community by paving the way for a unified, scalable, and interactive demonstration platform.

\section*{Limitations}

The focus of this paper was introducing an open online web application (demonstration system) to make the interaction of XNLP tasks available to as many practitioners as possible, but there are a couple of limitations
in the system and the model we proposed.
First, our system is based on the web service form, with the LLM running at the backend deployed at the online server, where sometimes when the Internet traffic is bad, the user may wait for too long to get the response.
Second, as the LLM essentially generates sequential texts of any inputs, there are chances that the output texts include problematic structured formatter (i.e., structural representations, cf. Figure \ref{fig:formatter}).
With ill-formed structural representations, it is problematic to parse them into correct data used for rendering into brat visualization, i.e., causing failure prediction.
Third, as one of the nature characteristics, LLM may sometimes generate false output, or do not obey the input instructions, which has been called the \emph{Hallucination} phenomenon \cite{abs-2307-03987}.
In such case, the user experience will be affected.
Lastly, the current version of the system is still at a basic stage, and there are functionalities at the user interface level that need further polishing and improvement in subsequent updates.

\section*{Ethics Statement}

Our XNLP system uses the LLM as backbone.
While the Vacuna model is fine-tuned on the pre-trained LLaMA model, which is known to contain some toxic contents \cite{SchickUS21}, an internal check does not reveal any toxic generation. 
However, there is a potential risk that the Vacuna could generate toxic text for users due to the underlying black-box LLM.

\bibliography{anthology}
\bibliographystyle{acl_natbib}

\appendix

\hfill
\section{Selection of Pre-defined XNLP Tasks}
\label{Selection of Pre-defined XNLP Tasks}

\begin{figure}[!h]
\centering
\includegraphics[width=0.98\columnwidth]{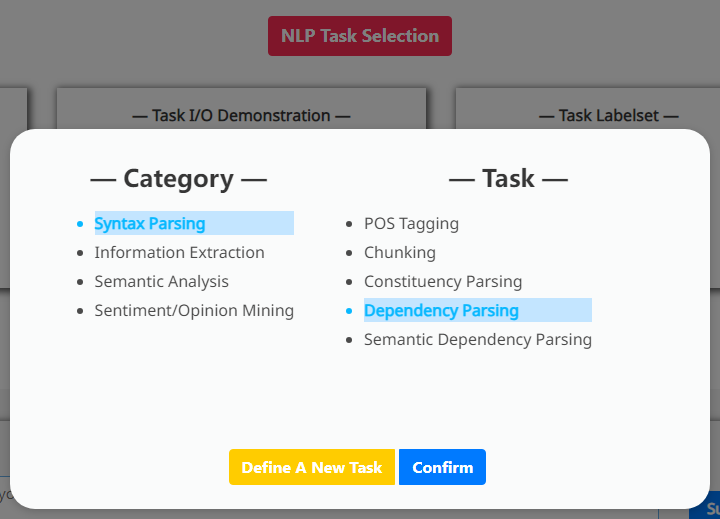}
\caption{
Screenshot of the selection panel of pre-defined XNLP tasks.
}
\label{task-selection}
\end{figure}

See Figure \ref{task-selection}.

\section{New XNLP Task Definition}
\label{New XNLP Task Definition}

\begin{figure*}[htbp]
  \centering
  \begin{subfigure}[b]{0.99\textwidth}
    \includegraphics[width=\textwidth]{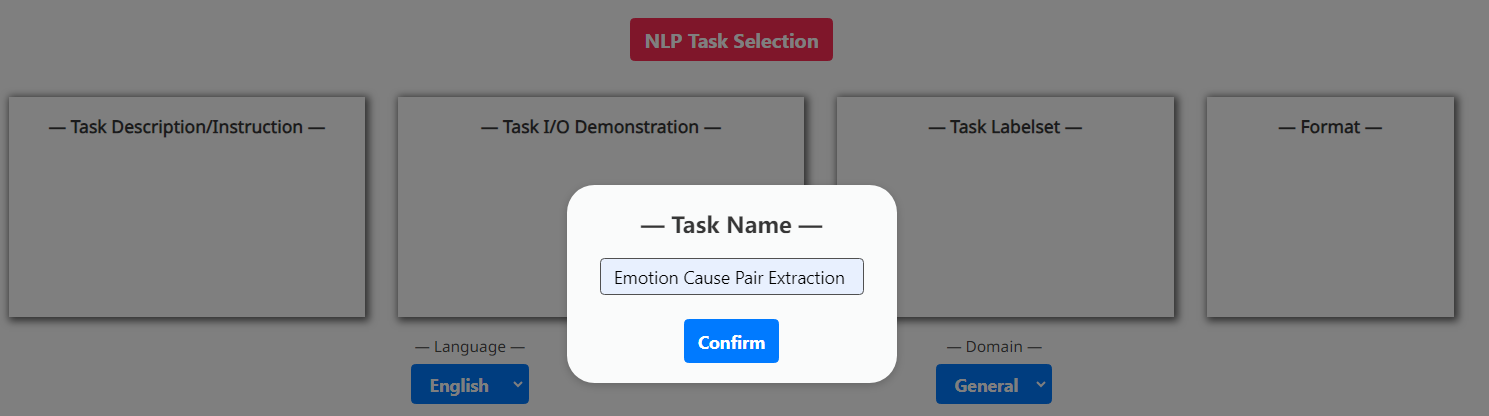}
    \caption{Step-1, name the task.}
  \end{subfigure}
  \begin{subfigure}[b]{0.99\textwidth}
    \includegraphics[width=\textwidth]{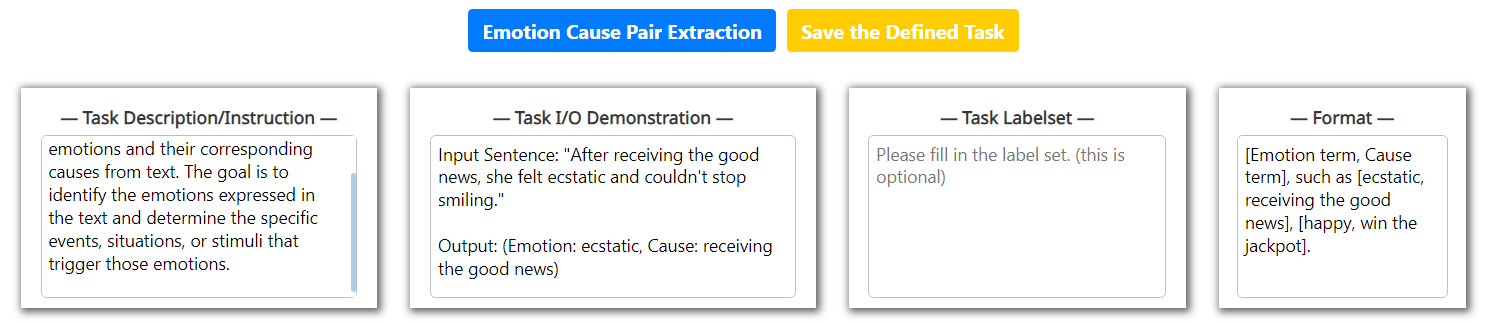}
    \caption{Step-2, fill in the task metadata.}
  \end{subfigure}
  \begin{subfigure}[b]{0.99\textwidth}
    \includegraphics[width=\textwidth]{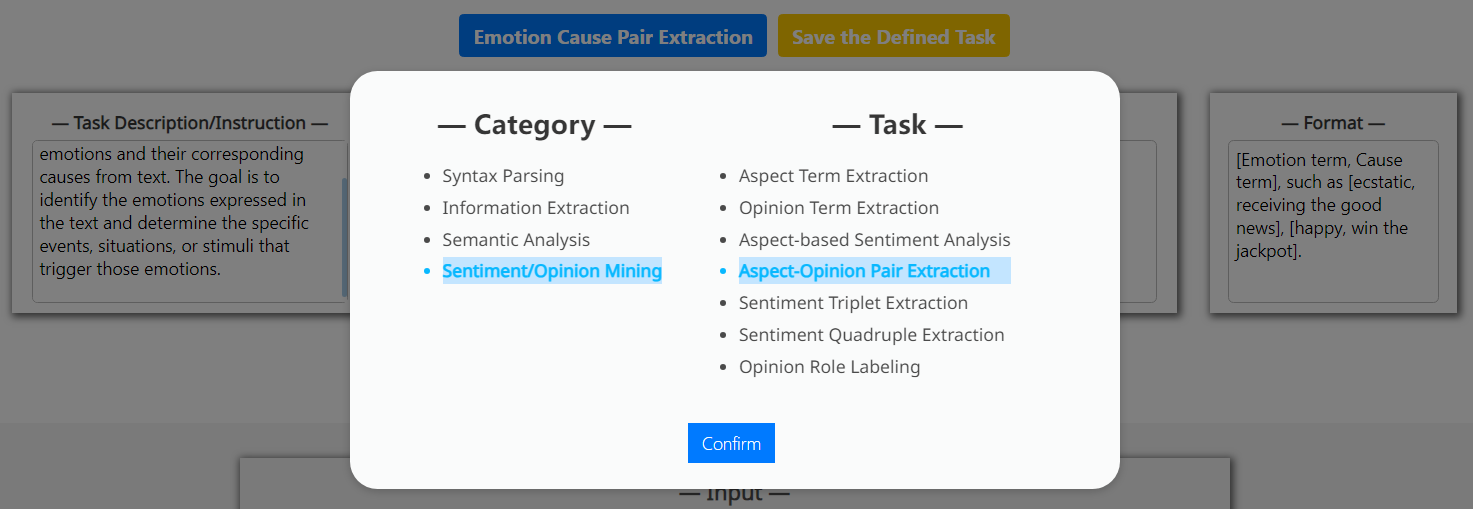}
    \caption{Step-3, select an executing format with a similar task,}
  \end{subfigure}
  \begin{subfigure}[b]{0.99\textwidth}
    \includegraphics[width=\textwidth]{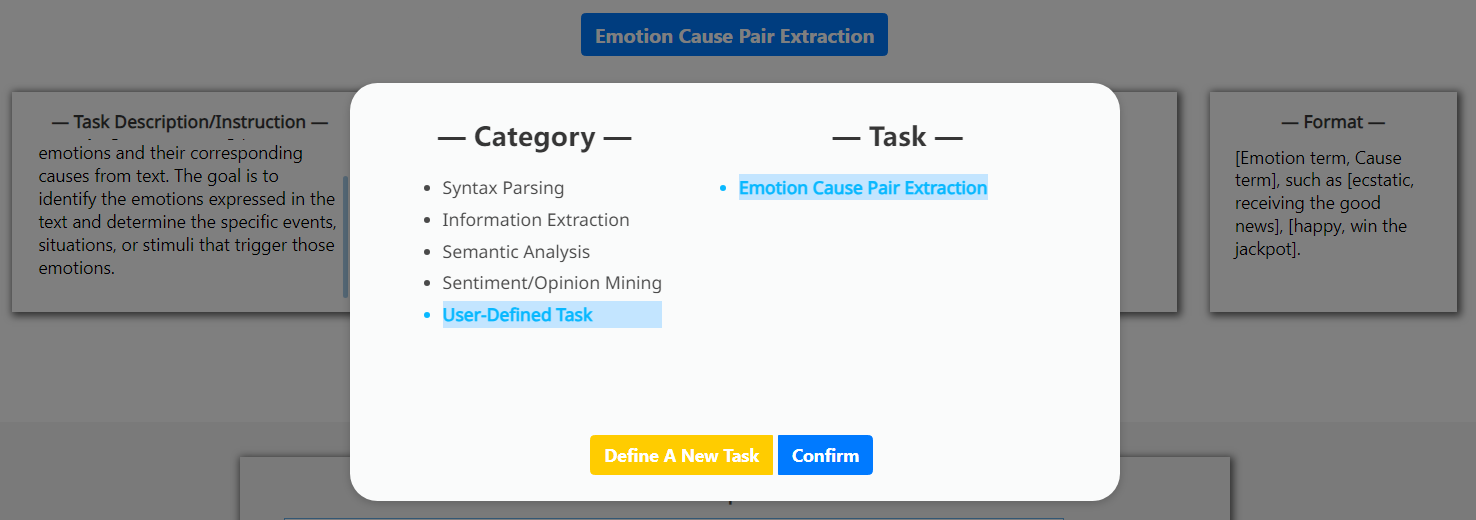}
    \caption{Step-4, confirm to define.}
  \end{subfigure}
  \caption{Screenshot of defining new XNLP task by the user.}
  \label{fig:definition}
\end{figure*}

See Figure \ref{fig:definition}.

\section{Multi-turn Interactions with User Feedback}
\label{Multi-turn Interactions with User Feedback}

\begin{figure*}[htbp]
  \centering
  \begin{subfigure}[b]{0.99\textwidth}
    \includegraphics[width=\textwidth]{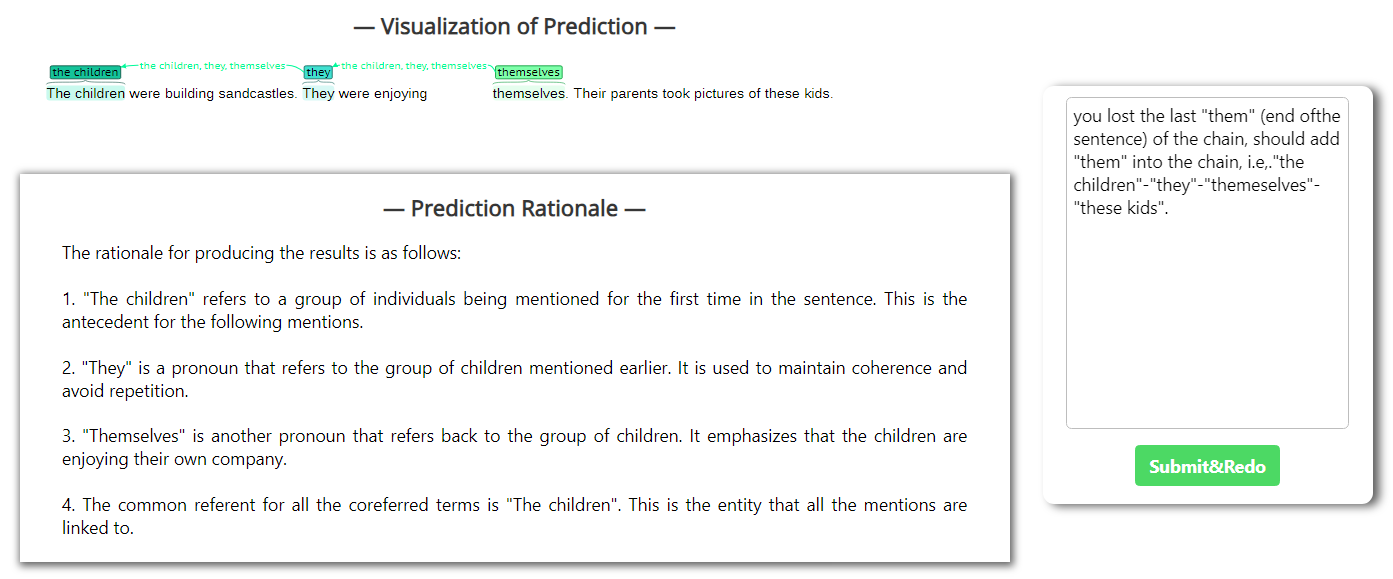}
    \caption{Before user feedback.}
  \end{subfigure}
  \hfill
  \begin{subfigure}[b]{0.99\textwidth}
    \includegraphics[width=\textwidth]{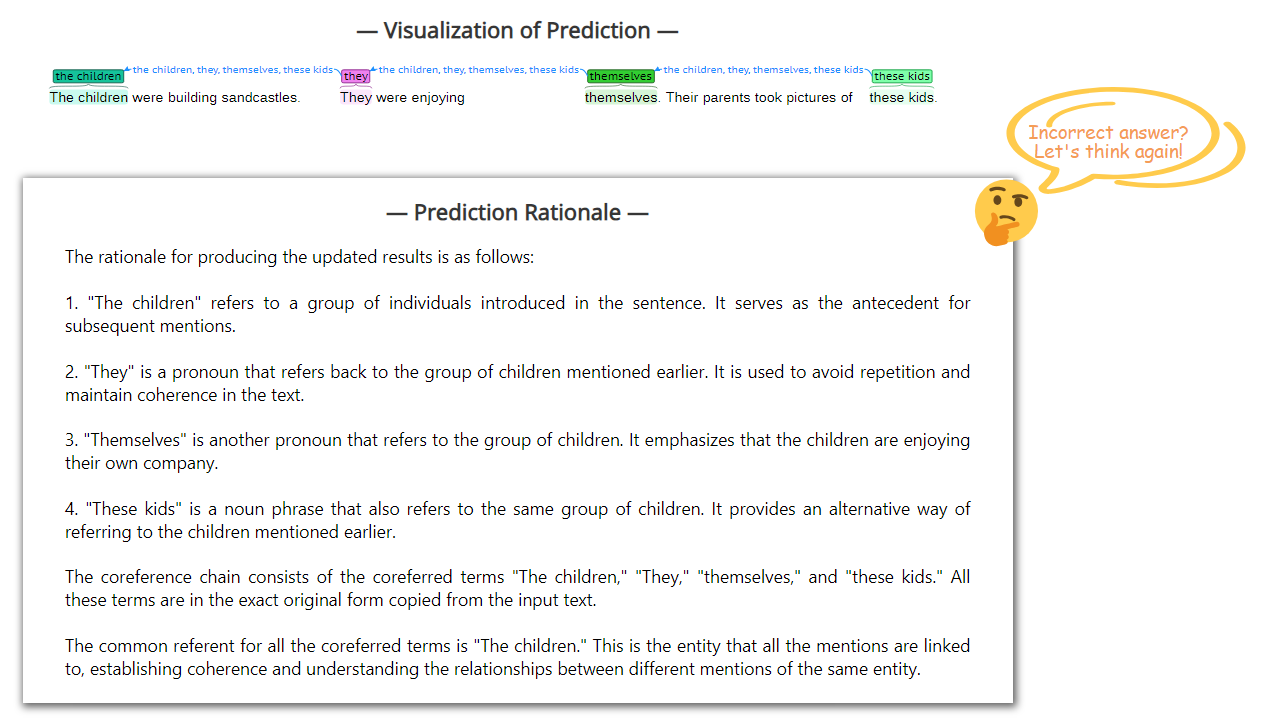}
    \caption{After user feedback.}
  \end{subfigure}
  \caption{System output before user feedback and after user feedback.}
  \label{fig:interaction}
\end{figure*}

See Figure \ref{fig:interaction}.

\begin{figure*}[!t]
\centering
\includegraphics[width=1\textwidth]{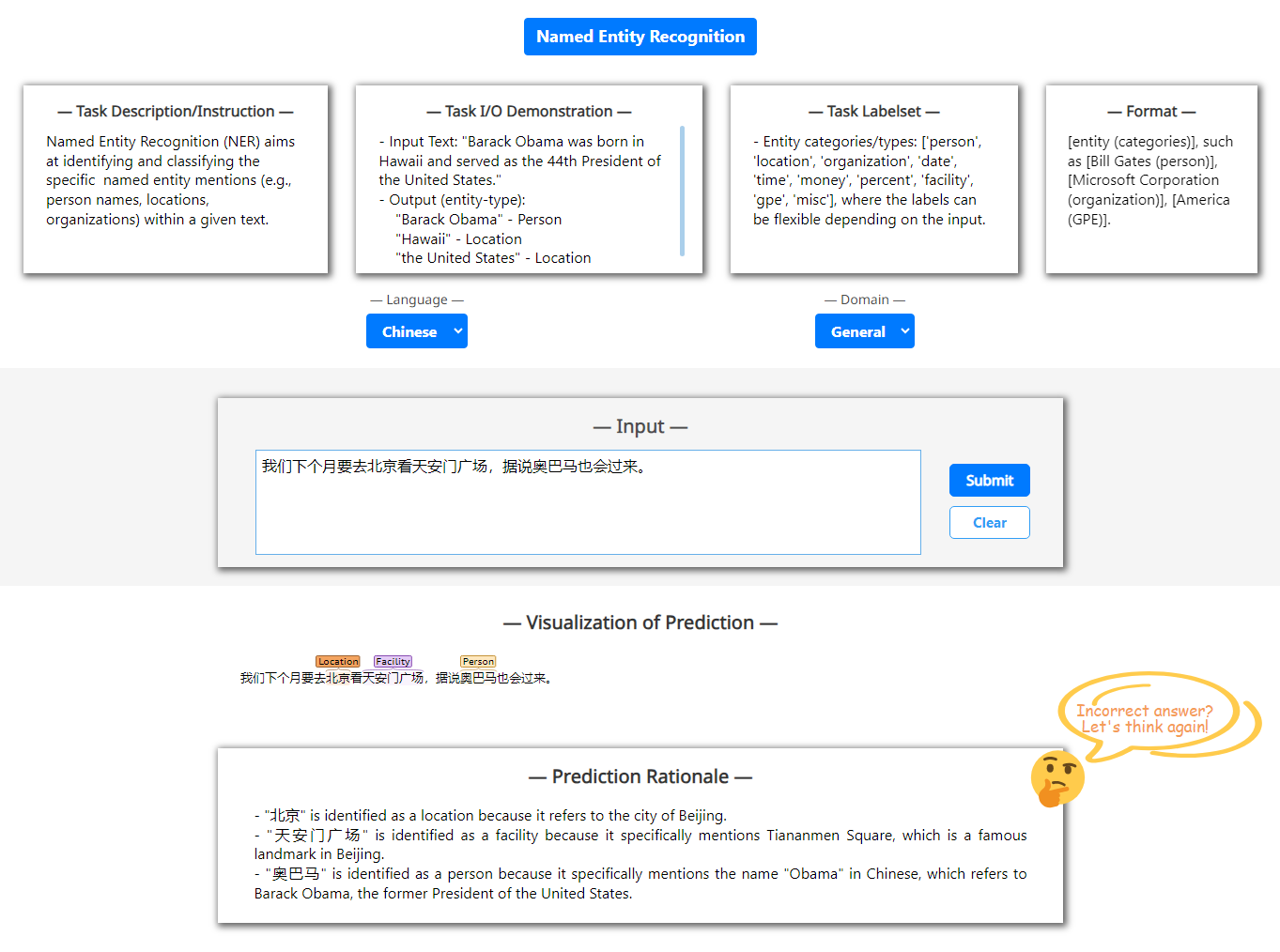}
\caption{
Input text in different languages (Chinese).
}
\label{fig:Chinese}
\end{figure*}

\begin{figure*}[!t]
\centering
\includegraphics[width=1\textwidth]{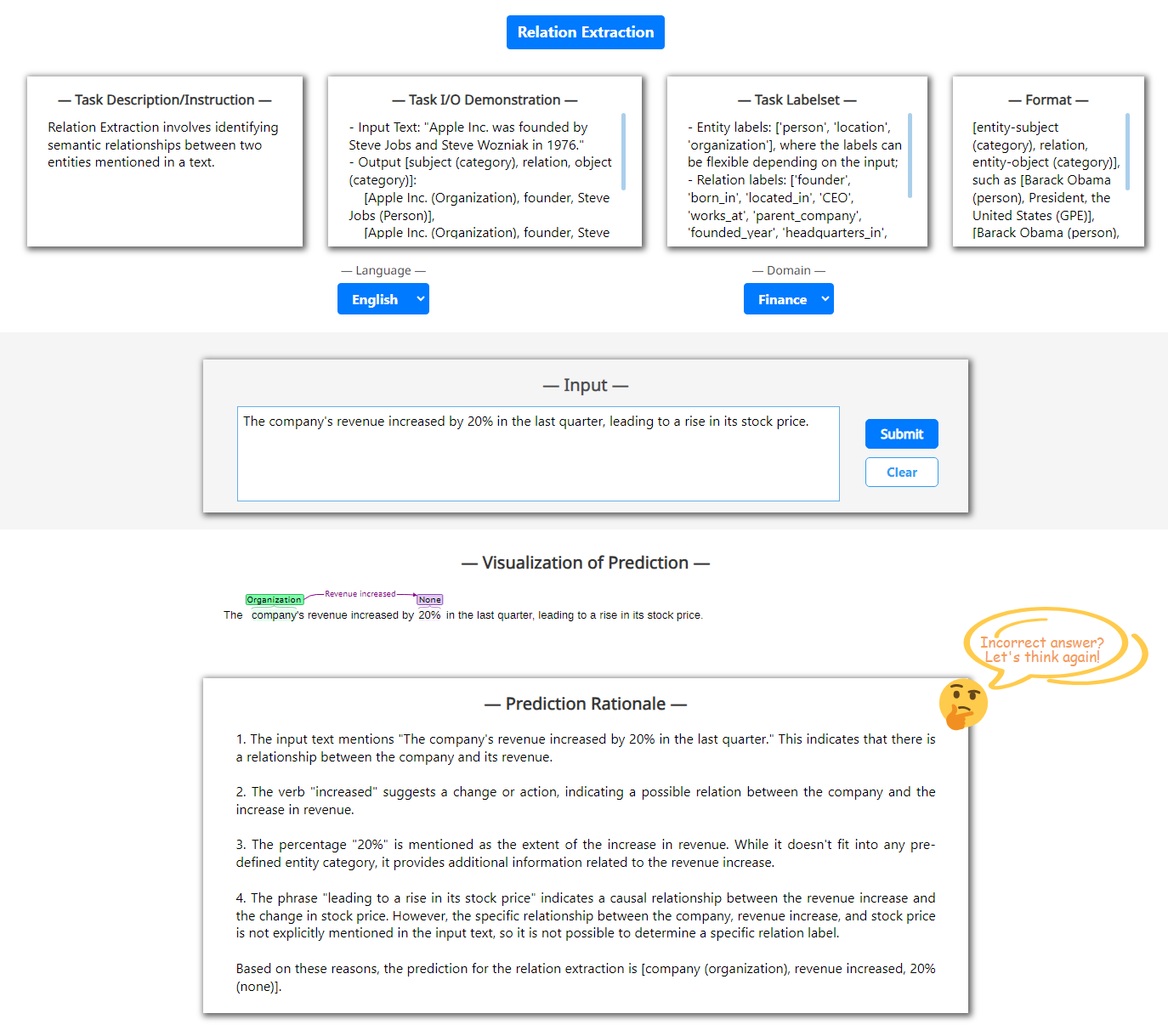}
\caption{
Input text in different domains (financial domain).
}
\label{fig:domain}
\end{figure*}

\section{Text in Different Language}
\label{Text in Different Language}

See Figure \ref{fig:Chinese}.

\section{Text in Different Domain}
\label{Text in Different Domain}

See Figure \ref{fig:domain}.

\end{document}